\documentclass{article} 
\usepackage{iclr2023_conference,times}

\usepackage{amsmath,amsfonts,bm}

\def\eqref#1{equation~\ref{#1}}

\def\1{\bm{1}}

\DeclareMathAlphabet{\mathsfit}{\encodingdefault}{\sfdefault}{m}{sl}
\SetMathAlphabet{\mathsfit}{bold}{\encodingdefault}{\sfdefault}{bx}{n}

\usepackage{graphicx,nicefrac}
\usepackage{tikz}
\usepackage{comment}
\usepackage{amsmath,amssymb} 
\usepackage{color}
\usepackage{booktabs}
\usepackage{listings}

\usepackage[hidelinks]{hyperref}
\usepackage[capitalize]{cleveref}

\crefname{section}{Sec.}{Secs.}
\Crefname{section}{Section}{Sections}
\Crefname{table}{Table}{Tables}
\crefname{table}{Tab.}{Tabs.}
\usepackage{wrapfig}
\usepackage{subcaption}
\usepackage{xspace}
\makeatletter
\DeclareRobustCommand\onedot{\futurelet\@let@token\@onedot}
\def\@onedot{\ifx\@let@token.\else.\null\fi\xspace}

\def\eg{\emph{e.g}\onedot} 
\def\ie{\emph{i.e}\onedot}

\makeatother
\usepackage{algorithm}
\usepackage{algorithmic}
\usepackage{epsfig}
\usepackage{verbatimbox}
\usepackage{amssymb}
\usepackage{pifont}
\usepackage{threeparttable}
\usepackage{multirow}
\usepackage{xr-hyper}
\usepackage{rotating}

\newcommand{\cmark}{\ding{51}\xspace}%
\newcommand{\xmark}{\ding{55}\xspace}%

\def \ours          {CPVT\xspace}
\def \oursbase      {CPVT-B\xspace}
\def \ourstiny      {CPVT-Ti\xspace}
\def \ourssmall     {CPVT-S\xspace}
\def \oursbasegap     {CPVT-B-GAP\xspace}
\def \ourssmallgap     {CPVT-S-GAP\xspace}
\def \ourstinygap      {CPVT-Ti-GAP\xspace}
\def  \peg {PEG\xspace}
\def  \pegs {PEGs\xspace}
\def \npe {DeiT w/o PE}

\def \pzo {\phantom{0}}

\usepackage{xfrac}
\usepackage[margin=1pt,font=normalsize,labelsep=period,tableposition=top]{caption} 

\def \alambic {\includegraphics[width=0.02\linewidth]{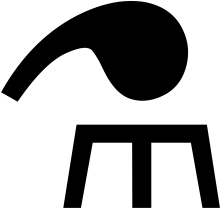}\xspace}

\makeatletter
\newcommand*{\addFileDependency}[1]{
  \typeout{(#1)}
  \@addtofilelist{#1}
  \IfFileExists{#1}{}{\typeout{No file #1.}}
}
\makeatother

\newcommand{\myparagraph}[1]{{\vspace{.5em} \noindent \bf #1}}

\usepackage{url}

\title{Conditional Positional Encodings for Vision Transformers}

\author{Xiangxiang Chu$^1$, ~ Zhi Tian$^1$, ~ Bo Zhang$^1$, ~ Xinlong Wang$^2$, ~ Chunhua Shen$^3$\thanks{Corresponding author.} \\
	$^1$ Meituan Inc. ~~ $^2$ Beijing Academy of AI ~~  $^3$ Zhejiang University, China \\
	{\tt\small \{chuxiangxiang, tianzhi02, zhangbo97\}@meituan.com}, \\
	{\tt\small xinlong.wang96@gmail.com, chunhua@me.com}
}
\iclrfinalcopy 
\begin{document}

\maketitle

\begin{abstract}
We propose a conditional positional encoding (CPE) scheme for vision Transformers~\citep{dosovitskiy2021an,touvron2020training}. Unlike previous fixed or learnable positional encodings that are predefined and independent of input tokens, CPE is dynamically generated and conditioned on the local neighborhood of the input tokens. As a result, CPE can easily generalize to the input sequences that are longer than what the model has ever seen during the training. Besides, CPE can keep the desired translation equivalence in vision tasks, resulting in improved performance. We implement CPE with a simple Position Encoding Generator (PEG) to get seamlessly incorporated into the current Transformer framework. Built on PEG, we present Conditional Position encoding Vision Transformer (CPVT). We demonstrate that CPVT has visually similar attention maps compared to those with learned positional encodings and delivers outperforming results.  Our
Code is available 
at: \href{https://github.com/Meituan-AutoML/CPVT}{\texttt{https://git.io/CPVT}}.
\end{abstract}

\section{Introduction}
Recently, Transformers~\citep{vaswani2017attention} have been viewed as a strong alternative to Convolutional Neural Networks (CNNs) in visual recognition tasks such as classification~\citep{dosovitskiy2021an} and detection~\citep{carion2020end,zhu2021deformable}. Unlike the convolution operation in CNNs, which has a limited  receptive field, the self-attention mechanism in the Transformers can capture the long-distance information and dynamically adapt the receptive field according to the image content. Consequently, Transformers are considered more flexible and powerful than CNNs, being promising to achieve more progress in visual recognition.

\begin{figure}[ht]
\centering
\begin{subfigure}{0.3\linewidth}
{
	\includegraphics[width=\columnwidth]{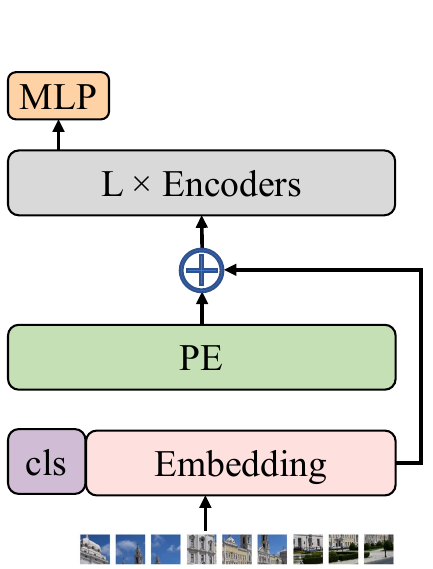}
	 \caption{ViT}
	 \label{subfig:vit}
}
\end{subfigure}
\begin{subfigure}{0.3\linewidth}
	\includegraphics[width=\columnwidth]{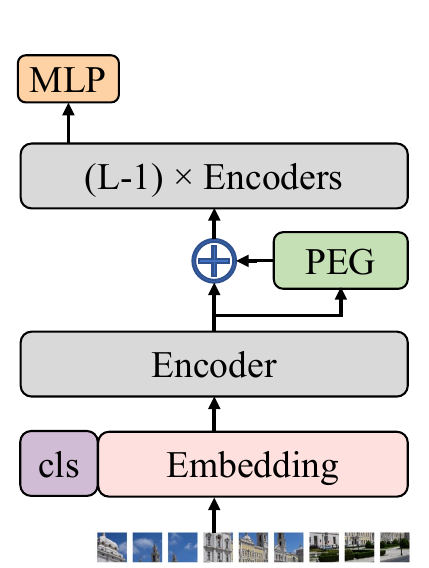}
	 \caption{CPVT}
	\label{subfig:cpvt}
\end{subfigure}
\begin{subfigure}{0.3\linewidth}
e    \includegraphics[width=\columnwidth]{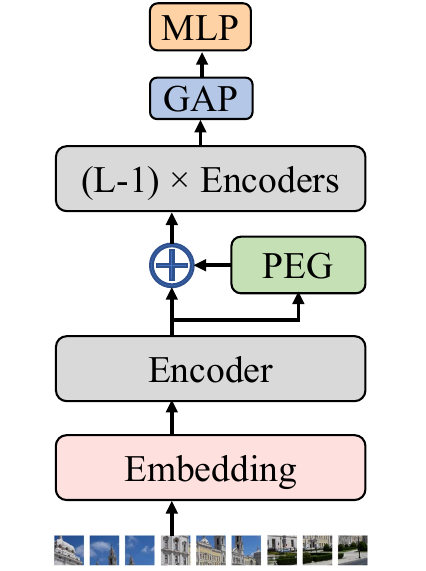}
    \caption{CPVT-GAP}
    \label{subfig:cpvt-gap}
\end{subfigure}
\caption{Vision Transformers: (a) ViT  \citep{dosovitskiy2021an} with explicit 1D learnable positional encodings (\texttt{PE}) (b) CPVT with conditional positional encoding from the proposed Position Encoding Generator (\texttt{PEG}) plugin, which is the \emph{default} choice. (c) CPVT-GAP without class token ({\tt cls}), but with global average pooling (GAP) over all items in the sequence. Note that GAP is a bonus version which has boosted performance.}
\label{fig:povt-schema}
\vskip -0.1in
\end{figure}

However, the self-attention operation in Transformers is permutation-invariant, which discards the order of the tokens in an input sequence. To mitigate this issue, previous works \citep{vaswani2017attention,dosovitskiy2021an} add the absolute positional encodings to each input token (see Figure~\ref{fig:povt-schema}a), which enables order-awareness. The positional encoding can either be learnable or fixed with sinusoidal functions of different frequencies. Despite being effective, these  positional encodings seriously harm the flexibility of the Transformers, hampering their broader applications. Taking the learnable version as an example, the encodings are often a vector of equal length to the input sequence, which are jointly updated with the network weights during training. As a result, the length and the value of the positional encodings are fixed once trained. During testing, it causes difficulties of handling the sequences longer than the ones in the training data.

The inability to adapt to longer input sequences during testing greatly limits the range of generalization. For instance, in vision tasks like object detection, we expect the model can be applied to the images of any size during inference, which might be much larger than the training images. A possible remedy is to use bicubic interpolation to upsample the positional encodings to the target length, but it degrades the performance without fine-tuning as later shown in our experiments. For vision in general, we expect that the models be translation-equivariant. For example, the output feature maps of CNNs  shift accordingly as the target objects are moved in the input images.
However, the absolute positional encoding scheme might break the translation equivalence because it adds unique positional encodings to each token (or each image patch). One may overcome the issue with relative positional encodings as in \citep{shaw2018self}. However, relative positional encodings not only come with extra computational costs, but also require modifying the implementation of the standard Transformers. Last but not least, the relative positional encodings cannot work equally well as the absolute ones, because the image recognition task still requires absolute position information \citep{islam2020much}, which the relative positional encodings fail to provide.

In this work, we advocate a novel positional encoding (PE) scheme to incorporate the position information into Transformers. Unlike the predefined and input-agnostic positional encodings used in previous works \citep{dosovitskiy2021an,vaswani2017attention,shaw2018self}, the proposed PE is dynamically generated and conditioned on the local neighborhood of input tokens. Thus, our positional encodings can change along with the input size and try to keep translation equivalence. We demonstrate that the vision transformers~\citep{dosovitskiy2021an,touvron2020training} with our new PE (\ie \ours, see Figure~\ref{subfig:cpvt-gap}) achieve even better performance. We summarize our contributions as,


\begin{itemize}
	\item We propose a novel positional encoding (PE) scheme, termed \emph{conditional position encodings} (CPE). CPE is dynamically generated with Positional Encoding Generators (\peg) and can be effortlessly implemented by the modern deep learning frameworks~\citep{paszke2019pytorch,abadi2016tensorflow,chen2015mxnet}, requiring no changes to the current Transformer APIs. Through an in-depth analysis and thorough experimentations, we unveil that this design affords both absolute and relative encoding yet it goes above and beyond.
	
	\item 
	 As opposed to widely-used absolute positional encodings, CPE can provide a kind of  stronger explicit bias towards the \textbf{translation equivalence} which is important to improve the performance of Transformers.  
	
	\item Built on CPE, we propose Conditional Position encoding Vision Transformer (CPVT). It achieves better performance than previous vison transformers~\citep{dosovitskiy2021an,touvron2020training}.
	
	\item CPE can well generalize to arbitrary input  resolutions, which are required in many important downstream tasks such as segmentation and detection. Through experiments we show that CPE can boost the segmentation and detection   performance for  pyramid transformers like \citep{wang2021pyramid}  by a clear margin.

	
\end{itemize}

\section{Related Work}

Since self-attention itself is permutation-equivariant (see \ref{sec:trans-equiv}),  positional encodings are commonly employed to incorporate the order of sequences \citep{vaswani2017attention}. The positional encodings can either be fixed or learnable, while either being absolute or relative. Vision transformers follow the same fashion to imbue the network with  positional information. 


\paragraph{Absolute Positional Encoding.} The absolute positional encoding is the most widely used. In the original transformer~\citep{vaswani2017attention}, the encodings are generated with the sinusoidal functions of different frequencies and then they are added to the inputs. Alternatively, the positional encodings can be learnable, where they are implemented with a fixed-dimension matrix/tensor and jointly updated with the model's parameters with SGD. 

\paragraph{Relative Positional Encoding.} The relative position encoding~\citep{shaw2018self} considers distances between the tokens in the input sequence. Compared to the absolute ones, the relative positional encodings can be translation-equivariant and can naturally handle the sequences longer than the longest sequences during training (\ie, being inductive). A 2-D relative position encoding is proposed for image classification in \citep{bello2019attention}, showing superiority to 2D sinusoidal embeddings. The relative positional encoding is further improved in XLNet~\citep{yang2019xlnet} and DeBERTa~\citep{he2020deberta}, showing better performance.

\paragraph{Other forms.} Complex-value embeddings \citep{wang2019encoding} are an extension to model global absolute encodings and show improvement.  RoFormer \citep{su2021roformer} utilizes a rotary position embedding to encode both absolute and relative position information for text classification. FLOATER \citep{liu2020learning} proposes a novel continuous dynamical model to capture position encodings. It is not limited by the maximum sequence length during training, meanwhile being parameter-efficient.

\paragraph{Similar designs to CPE.}  Convolutions are used to model local relations in ASR and machine translation \citep{gulati2020conformer,mohamed2019transformers,yang2019convolutional,yu2018qanet}. However, they are mainly limited to 1D signals. We instead process 2D vision images.

\section{Vision Transformer with Conditional Position Encodings}


\subsection{Motivation}\label{sec:mot}
In vision transformers, an input image of size $H\times W$ is split into patches with size $S \times S$, the number of patches is $N = \frac{HW}{S^2}$\footnote{$H$ and $W$ shall be divisible by $S$, respectively.}. The patches are added with the same number of learnable absolute positional encoding vectors. In this work, we argue that the positional encodings used here have two issues. First, it prevents the model from handling the sequences longer than the learnable PE. Second, it makes the model not translation-equivariant because a unique positional encoding vector is added to every one patch. The translation equivalence plays an important role in classification because we hope the networks' responses changes accordingly as the object moves in the image.

One may note that the first issue can be remedied by removing the positional encodings since except for the positional encodings, all other components (\eg, MHSA and FFN) of the vision transformer can directly be applied to longer sequences. However, this solution severely deteriorates the performance. This is understandable because the order of the input sequence is an important clue and the model has no way to extract the order without the positional encodings. The experiment results on ImageNet are shown in Table~\ref{fig: various_encoding}. By removing the positional encodings, DeiT-tiny's performance on ImageNet dramatically degrades from 72.2\% to 68.2\%.

\begin{table}[t]
	\caption{Comparison of various positional encoding (PE) strategies tested on ImageNet validation set in terms of the top-1 accuracy. Removing the positional encodings greatly damages the performance. The relative positional encodings have inferior performance to the absolute ones}\label{fig: various_encoding}
\vskip -0.1in
    \setlength{\tabcolsep}{1.5pt}
	\begin{center}
		\begin{tabular}{ l|c|c|c }
			\hline
			Model & Encoding  & Top-1@224(\%) &  Top-1@384(\%) \\
			\hline 
			DeiT-tiny~\citep{touvron2020training} & \xmark &68.2 & 68.6\\
			DeiT-tiny~\citep{touvron2020training} & learnable & 72.2 & 71.2\\
			DeiT-tiny~\citep{touvron2020training} & sin-cos& 72.3 & 70.8\\
			DeiT-tiny &  2D RPE~\citep{shaw2018self} & 70.5 & 69.8\\
			\hline
		\end{tabular}
	\end{center}
	\vskip -0.2in
\end{table}

Second, in DeiT~\citep{touvron2020training}, they show that we can interpolate the position encodings to make them have the same length of the longer sequences. However, this method requires fine-tuning the model a few more epochs, otherwise the performance will remarkably drop, as shown in Table~\ref{fig: various_encoding}. This goes contrary to what we would expect. With the higher-resolution inputs, we often expect a remarkable performance improvement without any fine-tuning. Finally, the relative position encodings~\citep{shaw2018self,bello2019attention} can cope with both the aforementioned issues. However, the relative positional encoding cannot provide \emph{absolute position information}, which is also important to the classification performance~\citep{islam2020much}. As shown in Table~\ref{fig: various_encoding}, the model with relative position encodings has inferior performance (70.5\% vs. 72.2\%).

\subsection{Conditional Positional Encodings}
We argue that a successful positional encoding for vision tasks should meet these requirements,
\begin{itemize}
\item[(1)] Making the input sequence \emph{permutation-variant} and providing  stronger explicit bias towards \emph{translation-equivariance}.
\item[(2)] Being inductive and able to handle the sequences longer than the ones during training.
\item[(3)] Having the ability to provide the absolute position to a certain degree. This is important to the performance as shown in ~\citep{islam2020much}.
\end{itemize}

In this work, we find that characterizing the local relationship by positional encodings is sufficient to meet all of the above. First, it is \emph{permutation-variant} because the permutation of input  sequences also affects the order in some local neighborhoods. However, translation of an object in an input image does not change the order in its local neighborhood, \ie, \emph{translation-equivariant} (see Section~\ref{sec:trans-equiv}). Second, the model can easily generalize to longer sequences since only the local neighborhoods of a token are involved. Besides, if the absolute position of any input token is known, the absolute position of all the other tokens can be inferred by the mutual relation between input tokens. We will show that the tokens on the borders can be aware of their absolute positions due to the commonly-used zero paddings.

\begin{wrapfigure}{r}{10cm}
\vspace{-15pt}
    \centering
	\includegraphics[width=0.7\textwidth]{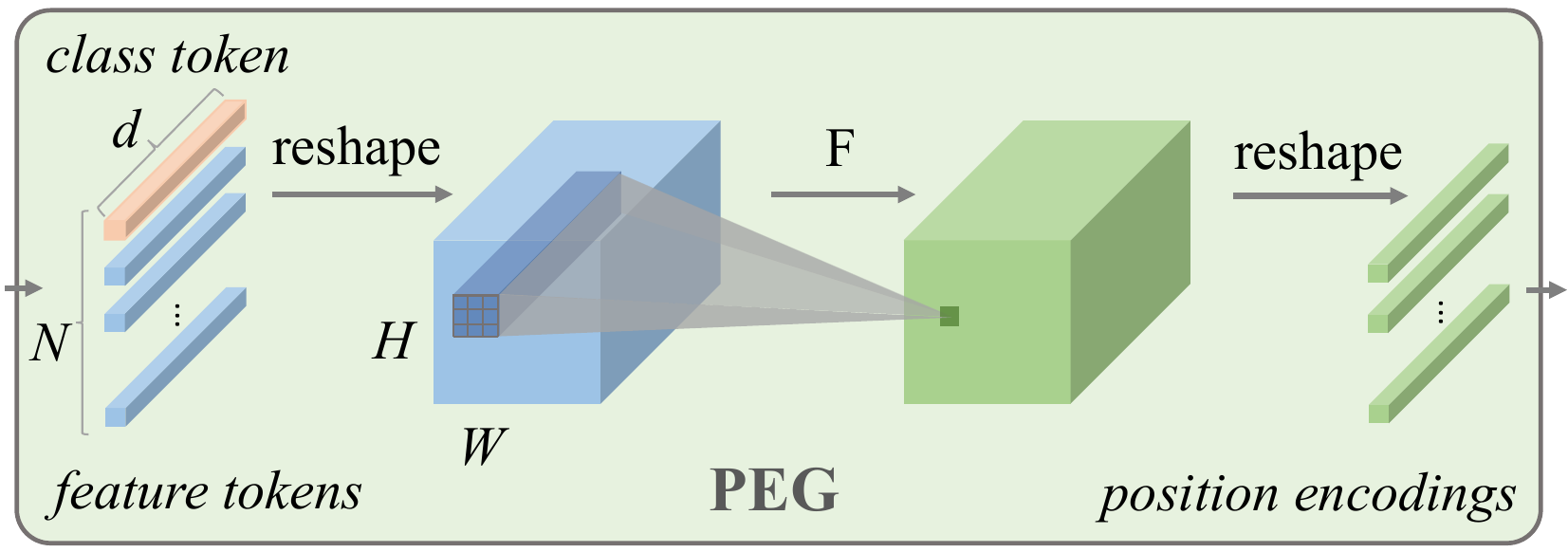}
	\caption{Schematic illustration of Positional Encoding Generator (\peg). Note $d$ is the embedding size, $N$ is the number of tokens. } 
	\label{fig:peg-scheme}
\end{wrapfigure}

Therefore, we propose \emph{positional encoding generators} (PEG) to dynamically produce the positional encodings conditioned on the local neighborhood of an input token.

\textbf{Positional Encoding Generator.} PEG is illustrated in Figure~\ref{fig:peg-scheme}. To condition on the local neighbors, we first reshape the flattened input sequence $ X \in \mathbb{R}^{B \times N\times C}$ of DeiT back to $X' \in \mathbb{R}^{B \times H \times W \times C}$ in the 2-D image space. Then, a function (denoted by $\mathcal{F}$ in Figure~\ref{fig:peg-scheme}) is repeatedly applied to the local patch in $X'$ to produce the conditional positional encodings $E^{B \times H \times W \times C}$. PEG can be efficiently implemented with a 2-D convolution with kernel $k$ ($k\geq3$) and $\frac{k-1}{2}$ zero paddings. Note that the zero paddings here are important to make the model be aware of the absolute positions, and $\mathcal{F}$ can be of various forms such as various types of convolutions and many others.

\subsection{Conditional Positional Encoding Vision Transformers}
Built on the conditional positional encodings, we propose our Conditional Positional Encoding Vision Transformers (\ours). Except that our positional encodings are conditional, we exactly follow ViT and DeiT to design our vision transformers and we also have three sizes \ours-Ti, \ours-S and \ours-B. Similar to the original positional encodings in DeiT, the conditional positional encodings are also added to the input sequence, as shown in Figure~\ref{fig:povt-schema} (b). In \ours, the position where PEG is applied is also important to the performance, which will be studied in the experiments.

In addition, both DeiT and ViT utilize an extra learnable class token to perform classification (\ie, \texttt{cls\_token} shown in Figure~\ref{fig:povt-schema} (a) and (b)). By design, the class token is not translation-invariant, although it can learn to be so. A simple alternative is to directly replace it with a global average pooling (GAP), which is inherently translation-invariant, resulting in our CVPT-GAP. Together with CPE, CVPT-GAP achieves much better image classification performance.

\section{Experiments}

\subsection{Setup}\label{subsec:setup}
\textbf{Datasets.} Following DeiT \citep{touvron2020training}, we use ILSVRC-2012 ImageNet dataset \citep{deng2009imagenet} with 1K classes and 1.3M images to train all our models. We report the results on the validation set with 50K images. Unlike ViT \citep{dosovitskiy2021an}, we do not use the much larger undisclosed JFT-300M dataset \citep{sun2017revisiting}.

\textbf{Model variants.} We have three models with various sizes to adapt to various computing scenarios. The detailed settings are shown in Table~\ref{tab:models_variants} (see \ref{app:arch-var}). All experiments in this paper are performed on Tesla V100 machines. Training the tiny model for 300 epochs takes about 1.3 days on a single node with 8 V100 GPU cards. \ourssmall and \oursbase take about 1.6 and 2.5 days, respectively. 

\textbf{Training details}
All the models (except for \oursbase) are trained for 300 epochs with a global batch size of 2048 on Tesla V100 machines using AdamW optimizer \citep{loshchilov2018decoupled}.  We do not tune the hyper-parameters and strictly comply with the  settings in DeiT \citep{touvron2020training}. The learning rate is scaled with this formula $lr_{\rm scale} = \nicefrac{0.0005 \cdot  {\rm Batch Size}_{\rm global}}{512}$. 
The detailed hyperparameters are in the \ref{app:hyp}.

\subsection{Generalization to Higher Resolutions}\label{subsec:scale}
As mentioned before, our proposed PEG can directly generalize to larger image sizes without any fine-tuning. We confirm this here by evaluating the models trained with $224\times224$ images on the $384\times384, 448\times448, 512\times512$  images, respectively. The results are shown in Table~\ref{tab: direct_scale}. With the $384\times384$ input images, the DeiT-tiny with learnable positional encodings degrades from 72.2\% to 71.2\%. When equipped with sine encoding, the tiny model degrades from 72.2\% to 70.8\%. In constrat, our \ours model with the proposed PEG can directly process the larger input images, and \ourstiny's performance is boosted from 73.4\% to 74.2\% when applied to $384\times384$ images. Our \ourstiny outperforms DeiT-tiny by 3.0\%. This gap continues to increase as the input resolution enlarges.

\begin{table}[ht]
	\caption{Direct evaluation on other resolutions  without fine-tuning. The models are trained on 224$\times$224. A simple PEG of a single layer of 3$\times$3 depth-wise convolution is used here}
	\label{tab: direct_scale}
	\vskip -0.1in
	\setlength{\tabcolsep}{3pt}
	\begin{center}
		\begin{tabular}{ l|c|c|c|c|c|c }
			\hline
			Model  & Params &160(\%) &224(\%) & 384(\%) &448(\%) & 512(\%)   \\
			\hline
			DeiT-tiny &6M& 65.6 & 72.2 &71.2 & 68.8 & 65.9\\
			DeiT-tiny (sin) &6M&65.2& 72.3 &70.8 &   68.2 & 65.1\\
			DeiT-tiny (no pos)& 6M&62.1& 68.2 &68.6 & 68.4  & 65.0 \\
			\ourstiny & 6M &66.8(+1.2)& 72.4(+0.2)&73.2(+2.0)& 71.8(+3.0)& 70.3(+4.4) \\
			\ourstiny $^\ddagger$  & 6M&67.7 (+2.1)& 73.4(+1.2)&74.2(+3.0)&72.6(+3.8) & 70.8(+4.9) \\
			\hline
			DeiT-small & 22M & 75.6& 79.9& 78.1 & 75.9& 72.6 \\
			\ourssmall & 22M &76.1(+0.5) & 79.9& 80.4(+1.5) & 78.6(+2.7)&76.8(+4.2)\\
			\hline
		    DeiT-base & 86M &79.1& 81.8& 79.7& 79.8& 78.2\\
		    \oursbase & 86M& 80.5(+1.4) &81.9(+0.1) & 82.3(+2.6) & 82.4(+2.6) & 81.0(+2.8)  \\
			\hline
		\end{tabular}
				\item $^\ddagger$: Insert one \peg each after the first encoder till the fifth encoder
	\end{center}
\end{table}



\subsection{\ours with Global Average Pooling}\label{subsec:gap}
By design, the proposed PEG is  translation-equivariant (ignore paddings). Thus, if we further use the translation-invariant global average pooling (GAP) instead of the \texttt{cls\_token} before the final classification layer of \ours. \ours can be translation-invariant, which should be beneficial to the ImageNet classification task. Note the using GAP here results in even less computation complexity because we do not need to compute the attention interaction between the class token and the image patches. As shown in Table~\ref{tab: GAP}, using GAP here can boost \ours by more than \textbf{1\%}. For example, equipping \ourstiny with GAP obtains 74.9\% top-1 accuracy on the ImageNet validation dataset, which outperforms DeiT-tiny by a large margin (+2.7\%). Moreover, it even exceeds DeiT-tiny model with distillation (74.5\%). In contrast,  DeiT with GAP cannot gain so much improvement (only 0.4\% as shown in Table~\ref{tab: GAP}) because the original learnable absolute PE is not translation-equivariant and thus GAP with the PE is not translation-invariant. Given the superior performance, we hope our model can be a strong PE alternative in vision transformers.

\begin{table}[ht]
	\caption{Performance comparison of Class Token (CLT) and global average pooling (GAP) on ImageNet. \ours's can be further boosted with GAP}
	\label{tab: GAP}
	\vskip -0.1in
    \setlength{\tabcolsep}{3pt}
	\begin{center}
		\begin{tabular}{ l|c|c|c|c }
			\hline
			 Model & Head & Params & Top-1 Acc& Top-5 Acc  \\
			 & & & (\%)  & (\%) \\
			\hline 
			DeiT-tiny \citep{touvron2020training} & CLT &6M& 72.2 &91.0  \\
			DeiT-tiny &GAP & 6M& 72.6 & 91.2  \\
			\ourstiny $^\ddagger$ & CLT  &  6M& 73.4& 91.8\\
			\ourstiny $^\ddagger$ & GAP  &  6M& \textbf{74.9}& \textbf{92.6}\\
			\hline
			 DeiT-small \citep{touvron2020training} &CLT& 22M & 79.9& 95.0 \\
			 DeiT-small  &GAP& 22M & 80.2& 95.2 \\
			 \ourssmall $^\ddagger$ &CLT & 23M & 80.5 & 95.2\\
			\ourssmall $^\ddagger$  &GAP& 23M & \textbf{81.5}& \textbf{95.7}\\
			\hline
		\end{tabular}
			\footnotesize
			\item $^\ddagger$: Insert one \peg each after the first encoder till the fifth encoder
	\end{center}
\end{table}

\subsection{Complexity of \peg}\label{sec:complexity}

\myparagraph{Few Parameters.} Given the model dimension $d$, the extra number of parameters introduced by \peg is $d \times l \times k^2$ if we choose  $l$ depth-wise convolutions with kernel $k$. If we use $l$ separable convolutions, this value becomes $l(d^2+k^2d)$. When $k=3$ and $l=1$, \ourstiny ($d=192$) brings about 1, 728 parameters. Note that DeiT-tiny utilizes learnable position encodings with $192\times14\times14=37632$ parameters. Therefore,  \ourstiny  has 35, 904 fewer number of parameters than DeiT-tiny. Even using 4 layers of separable convolutions, \ourstiny introduces only $38952 - 37632=960$ more parameters, which is negelectable compared to the 5.7M model parameters of DeiT-tiny.

\myparagraph{FLOPs.} As for FLOPs, $l$ layers of $k\times k$ depth-wise convolutions possesses $14 \times 14 \times d \times l \times k^2$ FLOPS. Taking the tiny model for example, it  involves $196\times 192\times9=0.34M$ FLOPS for the simple case $k=3$ and $l=1$, which is neglectable because the model has 2.1G FLOPs in total.

\subsection{Performance Comparison}\label{subsec:main-exp}

\begin{wrapfigure}{r}{7cm}
\vspace{-50pt}
\centering
	\includegraphics[width=0.5\textwidth]{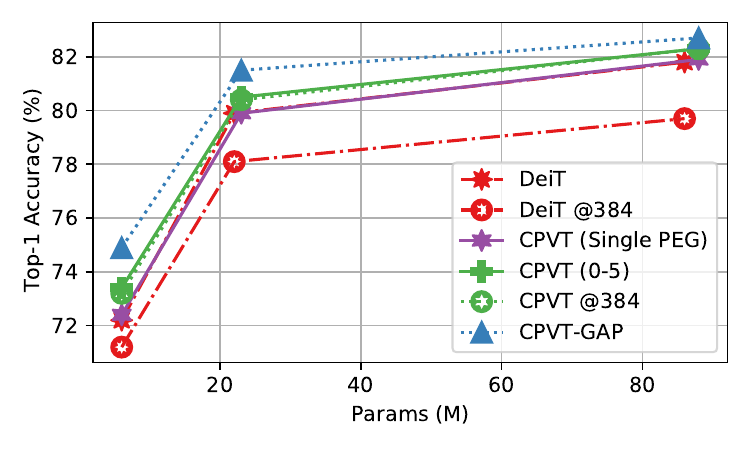}
	\caption{Comparison of CPVT and DeiT models under various configurations. Note CPVT@384 has improved performance. More PEGs can result in better performance. CPVT-GAP is the best.}
	\label{fig:cpvt-deit-comp}
	\vspace{-30pt}
\end{wrapfigure}

We evaluate the performance of \ours models on the ImageNet validation dataset and report the results in Table~\ref{tab:image_net}. Compared with DeiT, \emph{\ours models have much better top-1 accuracy with similar throughputs}. Our models can enjoy performance improvement when inputs are upscaled without fine-tuning, while DeiT degrades as discussed in Table
~\ref{tab: direct_scale},  see also Figure~\ref{fig:cpvt-deit-comp} for a clear comparison. Noticeably, Our model with GAP marked a new state-of-the-art for vision Transformers.

\begin{table}[ht]
	\caption{Comparison with ConvNets and Transformers on ImageNet and ImageNet Real \citep{beyer2020we}. \ours have much better performance compared with prior Transformers}
	\label{tab:image_net}
	\setlength{\tabcolsep}{2pt}
	\centering
	\begin{threeparttable}
		\begin{tabular}{@{\ }l |c|c@{\ }|c|c@{\ }|c@{\ }
  }
			\hline
			Models& Params(M)   &    Input      & throughput$^\star$ &ImNet & Real\\ 
			&&&&top-1 \%&top-1 \%\\
			\hline 
			\hline
			ResNet-50~\citep{he2016deep} & 25 & $224^{2}$ & 1226.1 & 76.2 &82.5 \\
			ResNet-101~\citep{he2016deep}& 45 & $224^{2}$ & \pzo753.6 & 77.4 & 83.7  \\
			ResNet-152~\citep{he2016deep} & 60 & $224^{2}$ & \pzo526.4 & 78.3 & 84.1\\
			
			RegNetY-4GF~\citep{radosavovic2020designing} &  21& $224^{2}$ & 1156.7 & 80.0 & 86.4 \\
			
			EfficientNet-B0~\citep{tan2019efficientnet} & 5 & $224^{2}$ & 2694.3 & 77.1&83.5\\
			EfficientNet-B1~\citep{tan2019efficientnet}& 8 & $240^{2}$ & 1662.5 & 79.1&84.9 \\
			EfficientNet-B2~\citep{tan2019efficientnet}& 9 & $260^{2}$ & 1255.7  & 80.1 &85.9 \\
			EfficientNet-B3~\citep{tan2019efficientnet}&12 &$300^{2}$  & \pzo732.1 & 81.6& 86.8 \\
			EfficientNet-B4~\citep{tan2019efficientnet} & 19 & $380^{2}$ & \pzo349.4 & 82.9 & 88.0 \\
			\hline
			ViT-B/16~\citep{dosovitskiy2021an}          & 86  & $384^{2}$ & \pzo\pzo85.9 & 77.9&-  \\
			ViT-L/16 & 307 & $384^{2}$ & \pzo\pzo27.3 & 76.5&- \\
			\hline
			DeiT-tiny w/o PE~\citep{touvron2020training} & 6 & $224^{2}$ & 2536.5 & 68.2 &- \\
			DeiT-tiny~\citep{touvron2020training} & 6 & $224^{2}$ & 2536.5 & 72.2& 80.1  \\
			DeiT-tiny (sine) & 6 & $224^{2}$ & 2536.5 & 72.3 & 80.3 \\
			\ourstiny$^\ddagger$ &6 & $224^{2}$& 2500.7 & 73.4& 81.3\\
			\textbf{\ourstinygap}$^\ddagger$ & 6 & $224^{2}$& 2520.1& \textbf{74.9}& 82.5\\
						\hline
			DeiT-tiny~\citep{touvron2020training}\alambic & 6 & $224^{2}$ & 2536.5& 74.5&82.1 \\
			\textbf{\ourstiny}\alambic  &6 & $224^{2}$ &2500.7& \textbf{75.9} & 83.0 \\
						
			\hline
			DeiT-small~\citep{touvron2020training}  & 22 & $224^{2}$ & \pzo940.4 & 79.9& 85.7 \\
			\ourssmall$^\ddagger$ & 23  & $224^{2}$ & \pzo 930.5 & 80.5 & 86.0  \\
			\textbf{\ourssmallgap}$^\ddagger$  & 23  & $224^{2}$ & \pzo 942.3 & \textbf{81.5} &86.6 \\
			\hline
			DeiT-base~\citep{touvron2020training}  & 86& $224^{2}$ & \pzo292.3 & 81.8 & 86.7\\
			\oursbase $^\ddagger$ & 88 & $224^{2}$ & \pzo285.5 & 82.3 &87.0 \\
			\textbf{\oursbasegap}$^\ddagger$ & 88 &$224^{2}$ & \pzo290.2 & \textbf{82.7} &87.7 \\

			\hline
		\end{tabular}
		\begin{tablenotes}
			\footnotesize
			\item $^\star$: Measured in \emph{img/s}  on a 16GB V100 GPU as in \citep{touvron2020training}. 
			\item $^\ddagger$: Insert one \peg each after the first encoder till the fifth encoder
			\item \alambic : trained with hard distillation using RegNetY-160 as the teacher.
		\end{tablenotes}
	\end{threeparttable}
\end{table}

	
We further train \ourstiny and  DeiT-tiny using the aforementioned  training settings plus the hard distillation proposed in \citep{touvron2020training}. Specifically, we use RegNetY-160 \citep{radosavovic2020designing} as the teacher. \ours obtains 75.9\%, exceeding DeiT-tiny by 1.4\%.

\subsection{PEG on Pyramid Transformer Architectures}
PVT \citep{wang2021pyramid} is a
vision transformer with the multi-stage design like ResNet \citep{he2016deep}. Swin \citep{liu2021swin} is a follow-up work and comes with higher performance. We apply our method on both to demonstrate its generalization ability. 
\paragraph{ImageNet classification.}
Specifically, we remove its learnable PE and apply our PEG in position 0 of each stage with a GAP head. We use the same training settings to make a fair comparison and show the results in Table~\ref{tab: PVT_peg}. Our method can significantly boost PVT-tiny by 3.1\% and Swin-tiny by 1.15\% on ImageNet (c.f. \ref{app:swin-tiny-peg}). We also evaluate the performance of PEG on some downstream semantic segmentation and object detection tasks (see \ref{app:seg-det}). Note these tasks usually handle the various input resolutions as the training because multi-scale data augmentation is extensively used.

\section{Ablation Study}

\subsection{Positional encoding or merely a hybrid?}

One might suspect that the PEG's improvement comes from the extra \emph{learnable parameters} introduced by the convolutional layers in PEG, instead of the local relationship retained by PEG. 
One way to test the function of PEG is only adding it when calculating Q and K in the attention layer, so that only the positional information of PEG is passed through. We can achieve 71.3\% top-1 accuracy on ImageNet with DeiT-tiny. This is significantly better than DeiT-tiny w/o PE (68.2\%) and is similar to the one with PEG on Q, K and V (72.4\%), which suggests that PEG mainly serves as a positional encoding scheme.

\begin{wraptable}{r}{7cm}
\vspace{-15pt}
	\caption{Positional encoding rather than added parameters gives the most improvement}
	\label{tab: learnable_not}
    	\setlength{\tabcolsep}{3pt}
	\begin{center}		
		\begin{tabular}{|c|c|c|c|c|}
			\hline
			Kernel &Style & Params & Top-1 Acc  \\
			& & (M) & (\%) \\
			\hline 
			none & - &5.68& 68.2  \\			
			3  & fixed (random init)&5.68 & 71.3\\
			3 & fixed (learned init) & 5.68 & 72.3 \\
			1 (12 $\times$) & learnable & 6.13 & 68.6\\
			3 & learnable & 5.68 & \textbf{72.4}\\
			\hline
		\end{tabular}
	\end{center}
\end{wraptable}

We also design another experiment to remove this concern. By randomly-initializing a 3$\times$3 PEG and fixing its weights during the training, we can obtain 71.3\% accuracy (Table~\ref{tab: learnable_not}), which is much higher (3.1\%$\uparrow$) than DeiT without any PE (68.2\%). Since the weights of PEG are fixed and the performance improvement can only be due to the introduced position information. On the contrary, when we exhaustively use 12 convolutional layers (kernel size being 1, \ie, not producing local relationship) to replace the PEG, these layers have much more learnable parameters than PEG. However, it only boosts the performance by 0.4\% to 68.6\%.

Another interesting finding is that fixing a learned PEG also helps training. When we initialize with a learned PEG instead of the random values and train the tiny version of the model from scratch while keeping the PEG fixed, the model can also achieve 72.3\% top-1 accuracy on ImageNet. This is very close to the learnable PEG (72.4\%). 

\subsection{PEG Position in \ours}
We also experiment by varying the position of the PEG in the model. Table~\ref{tab: position} (left) presents the ablations for variable positions (denoted as PosIdx) based on the tiny model. \emph{We consider the input of the first encoder by index -1.} Therefore, position 0 is the output of the first encoder block. PEG shows strong performance ($\sim$72.4\%) when it is placed at [0, 3].

Note that positioning the PEG at 0 can have much better performance than positioning it at -1 (\ie, before the first encoder), as shown in Table~\ref{tab: position} (left). We observe that the difference between the two situations is they have different receptive fields. Specifically, the former has a global field while the latter can only see a local area. Hence, \emph{they are supposed to work similarly well if we enlarge the convolution's kernel size}. To verify our hypothesis, we use a quite large kernel size 27 with a padding size 13 at position -1, whose result is reported in Table~\ref{tab: position} (right).  It achieves similar performance to the one positioning the PEG at 0 (72.5\%), which verifies our assumption.

\begin{table}[ht]
	\caption{Comparison of different plugin positions (left) and kernels (right) using DeiT-tiny}
	\label{tab: position}
	\begin{center}
		\begin{tabular}{|r|c|c|}
			\hline
			PosIdx &  Top-1 (\%) & Top-5 (\%) \\
			\hline 
			none & 68.2 & 88.7 \\
			$-1$   & 70.6& 90.2\\
		        0 & \textbf{72.4} & \textbf{91.2} \\
		        3 & 72.3 & 91.1 \\
			6  & 71.7 & 90.8\\
			10 & 69.0& 89.1 \\
			\hline
		\end{tabular}
		\hspace{2pt}
		\begin{tabular}{|c|c|c|c|c|}
			\hline
			PosIdx & kernel &Params& Top-1 (\%) & Top-5 (\%) \\
			\hline 
			-1  & 3$\times$3&5.7M&70.6&  90.2\\
			-1  & 27$\times$27 &5.8M&\textbf{72.5}& \textbf{91.3}\\
			\hline
		\end{tabular}
	\end{center}
\end{table}


\subsection{Comparisons with other positional encodings}\label{subsec:enc-type}
We compare PEG with other commonly used encodings: absolute positional encoding (e.g. sinusoidal \citep{vaswani2017attention}),  \emph{relative positional encoding} (RPE) \citep{shaw2018self} and \emph{learnable encoding} (LE) \citep{devlin2019bert,radford2018improving}, as shown in Table~\ref{tab: compare_encoding}. 

\begin{table}[t]
\vspace{-10pt}
	\caption{Comparison of various positional encoding strategies. LE: learnable positional encoding. RPE: relative positional encoding}
	\label{tab: compare_encoding}
    \setlength{\tabcolsep}{2pt}
	\begin{center}
		\begin{tabular}{|l|c|c|c|c|}
			\hline
			Model & \peg Pos & Encoding  & Top-1 & Top-5\\
			& & &  (\%)  & (\%)  \\
			\hline 
			DeiT-tiny~(\citeyear{touvron2020training}) & - & LE & 72.2 & 91.0\\
			DeiT-tiny & - & 2D sin-cos& 72.3 & 91.0\\
			DeiT-tiny & - & 2D RPE & 70.5 & 90.0\\
			\ourstiny & 0-1 & \peg& 72.4 & 91.2\\
			\ourstiny  & 0-1 &\peg+ LE&72.9& 91.4 \\
			\ourstiny  & 0-1 &4$\times$\peg+ LE&72.9& 91.4 \\
			\textbf{\ourstiny} & 0-5 & \peg & \textbf{73.4} & \textbf{91.8} \\
			\hline
		\end{tabular}
	\end{center}
\end{table}


DeiT-tiny obtains 72.2\% with the learnable absolute PE. We experiment with the 2-D sinusoidal encodings and it achieves on-par performance. For RPE, we follow \citep{shaw2018self} and set the local range hyper-parameter $K$ as 8,  with which we obtain 70.5\%. RPE here does not encode any absolute position information, see discussion in \ref{app:rpe-vs-pe} and \ref{subsec:zero-pad}.

Moreover, we combine the learnable absolute PE with a single-layer PEG. This boosts the baseline \ourstiny (0-1) by 0.5\%. If we use 4-layer PEG, it can achieve 72.9\%. If we add a PEG to each of the first five blocks,  we can obtain 73.4\%, which is better than stacking them within one block.

\textbf{CPE is  not  a simple combination of APE and RPE. }   We further compare our method with a baseline with combination of APE and RPE.  Specifically, we use learnable positional encoding (LE) as DeiT at the beginning of the model and supply 2D RPE for every transformer block. This setting achieves 72.4\% top-1 accuracy on ImageNet, which is comparable to a single PEG (72.4\%). Nevertheless, this experiment does not necessarily indicate that our CPE is a simple combination of APE and RPE. When tested on different resolutions, this baseline cannot scale well compared to ours (Table~\ref{tab:APE_RPE_combine}). RPE is not able to adequately mitigate the performance degradation on top of LE. This shall be seen as a major difference. 

\begin{table}[ht]
	\caption{Direct evaluation on other resolutions  without fine-tuning. The models are trained on 224$\times$224. CPE outperforms LE+RPE combination on untrained resolutions.}
	\label{tab:APE_RPE_combine}
	\setlength{\tabcolsep}{3pt}
	\begin{center}
		\begin{tabular}{|l|c|c|c|c|c|c|}
			\hline
			Model  & Positional Params  &160(\%) &224(\%) & 384(\%) &448(\%) & 512(\%)   \\
			\hline
			DeiT-tiny (LE+RPE) &40011& 65.6 & 72.4 &70.8 & 68.4 & 65.6\\
			DeiT-tiny (PEG at Pos 0)&1920&66.8& 72.4 &73.2 &   71.8& 70.3\\
			\hline
		\end{tabular}
	\end{center}
\end{table}

\textbf{PEG can continuously improve the performance if stacked more.} We use LE not only at the beginning but also in the next 5 layers to have a similar thing as 0-5 PEG configuration.This setting achieves 72.7\% top-1 accuracy on ImageNet, which is 0.7\% lower than PEG (0-5). This setting suggests that it is also beneficial to have more of LEs, but not as good as ours. It is expected since we exploit relative information via PEGs at the same time.

\section{Conclusion}
We introduced \ours, a novel method to provide the position information in vision transformers, which dynamically generates the position encodings based on the local neighbors of each input token. Through extensive experimental studies, we demonstrate that our proposed positional encodings can achieve stronger performance than the previous positional encodings. The transformer models with our positional encodings can naturally process longer input sequences and keep the desired translation equivalence in vision tasks. Moreover, our positional encodings are easy to implement and come with negligible cost. We look forward to a broader application of our  method in transformer-driven vision tasks like segmentation and video processing.

\bibliography{egbib}
\bibliographystyle{iclr2023_conference}

\newpage
\appendix

\section{Translation Equivariance}\label{sec:trans-equiv}

The term translation-equivariance  means the output feature maps can be equally translated with the input signal. Imagine there is a person in the left-top of an image, if the person is moved to the right-bottom, the output feature maps will change accordingly. This property is very important to the success of convolution network. Convolution (ignoring paddings), RPE, and self-attention are all translation-equivariant operations (regardless of their receptive field). It's nontrivial to make  absolute positional encodings like DeiT (using learnable positional encoding)  translation-equivariant since different absolute positions will be added if the input signal is translated. Note that our method is not strictly translation-equivariant because of the zero padding. Instead, it provides a kind of stronger explicit bias towards the translation-equivariant property.

\section{Experiment Details}

\subsection{Architecture variants of \ours}\label{app:arch-var}
\begin{table}[ht]
	\caption{\ours architecture variants.
	The larger model, \oursbase, has the same architecture as ViT-B~\citep{dosovitskiy2021an} and DeiT-B \citep{touvron2020training}. \ourssmall and \ourstiny have the same architecture as DeiT-small and DeiT-tiny respectively}\label{tab:models_variants}
\setlength{\tabcolsep}{3pt}
	\begin{center}
	\small
	\begin{tabular}{l|c|c|c|c}
	\hline
	Model               & \#channels & \#heads & \#layers & \#params  \\
	\hline
	\ourstiny    &  \pzo192  & \pzo3 & 12 & \pzo\pzo6M \\
	\ourssmall    &  \pzo384 & \pzo6 & 12 & \pzo22M \\
	\oursbase    & \pzo768   & 12 & 12 & \pzo86M \\
	\hline
	\end{tabular}
	\end{center}
\end{table}

\subsection{The Hyperparameters of \ours}\label{app:hyp}

As for the  ImageNet classification task, we use exactly the same hyperparameters as DeiT except for the base  model because it is not always stably trained using AdamW. The  detailed setting is shown in Table~\ref{tab:comp_hyperparameters}.

\begin{table}[h]
	\caption{Hyper-parameters for  ViT, DeiT and \ours}
	\label{tab:comp_hyperparameters}
	\centering
	\scalebox{0.9}
	{
		\begin{tabular}{l|c|c|c}
			\hline
			Methods & ViT  & DeiT & \ours \\

			\hline\hline
			Epochs   & 300 & 300  &300      \\ 

			Batch size & 4096 & 1024& 1024\\
			Optimizer & AdamW & AdamW&LAMB\\
			Learning rate decay & cosine & cosine & cosine \\
			Weight decay        & 0.3    & 0.05  & 0.05  \\
			Warmup epochs  & 3.4 & 5   & 5     \\
			Label smoothing $\varepsilon$ \citep{szegedy2016rethinking}  & \xmark & 0.1  & 0.1   \\
			Dropout \citep{srivastava2014dropout}      & 0.1 & \xmark & \xmark    \\
			Stoch. Depth \citep{huang2016deep} & \xmark & 0.1 & 0.1\\
			Repeated Aug \citep{hoffer2020augment} & \xmark & \cmark  & \cmark \\
			Gradient Clip. & \cmark & \xmark  & \xmark\\
			Rand Augment \citep{cubuk2020randaugment}  & \xmark        & 9/0.5  & 9/0.5 \\
			Mixup prob. \citep{zhang2018mixup}  & \xmark & 0.8   & 0.8    \\
			Cutmix prob. \citep{yun2019cutmix}   & \xmark & 1.0     & 1.0 \\
			Erasing prob. \citep{zhong2020random}    & \xmark & 0.25   & 0.25  \\
			\hline
		\end{tabular}
	}
\end{table}

\subsection{Importance of Zero Paddings}\label{subsec:zero-pad}
We design an experiment to verify the importance of the \emph{zero paddings}, which can help the model infer the absolute positional information. Specifically, we use \ourssmall and simply remove the zero paddings from \ours while keeping all other settings unchanged. Table~\ref{tab: padding} shows that this can only obtain 70.5\%, which indicates that the zero paddings and absolute positional information play important roles in classifying objects. 

\begin{table}[ht]
	\caption{Ablation study on ImageNet performance w/ or w/o zero paddings}
	\label{tab: padding}
	\begin{center}
		\small
		\begin{tabular}{l|c|c|c}
			\hline
			Model & Padding  & Top-1 Acc(\%) & Top-5 Acc(\%) \\
			\hline 
			\multirow{2}{*}{\ourstiny}  & \checkmark & \textbf{72.4}& \textbf{91.2}\\
			& \xmark& 70.5 & 89.8\\
			\hline
		\end{tabular}
	\end{center}
\end{table}

\subsection{Single PEG vs. Multiple PEGs}
We further evaluate whether or not using \emph{multi-position} encodings can benefit the performance in Table~\ref{tab: ablation_Number_of_Plug-in}. Notice we denote by $i$-$j$ the inserted positions of \peg which start from the $i$-th encoder and end at the $j-1$-th one (inclusion). By inserting \pegs to five positions, the top-1 accuracy of the tiny model can achieve 73.4\%, which surpasses  DeiT-tiny by 1.2\%.  Similarly, \ourssmall can achieve 80.5\%. It turns out more PEGs do help, but up to a level where more PEGs become incremental (0-5 vs. 0-11).

\begin{table}[ht]
	\caption{\ours's sensitivity to number of plugin positions}
	\label{tab: ablation_Number_of_Plug-in}
	\vskip -1in
    	\setlength{\tabcolsep}{3pt}
	\begin{center}		
	\small
		\begin{tabular}{r|c|r|c|c}
			\hline
			Positions & Model & Params& Top-1 Acc & Top-5 Acc \\
			& &(M) &(\%) & (\%) \\
			\hline 
			0-1 & tiny &5.7& 72.4 &91.2  \\
			0-5 & tiny & 5.9& \textbf{73.4} & \textbf{91.8}  \\
			0-11 & tiny  &  6.1 & \textbf{73.4}& \textbf{91.8}\\
			\hline
			0-1 & small & 22.0 & 79.9& 95.0 \\
			0-5 & small & 22.9 & 80.5 & \textbf{95.2}\\
			0-11 & small & 23.8 & \textbf{80.6} & \textbf{95.2}\\
			\hline
		\end{tabular}
	\end{center}
\end{table}

\subsection{Classfication Evaluation of Swin with PEG}\label{app:swin-tiny-peg}

\begin{figure}[ht]
\centering
\includegraphics[width=0.605\columnwidth]{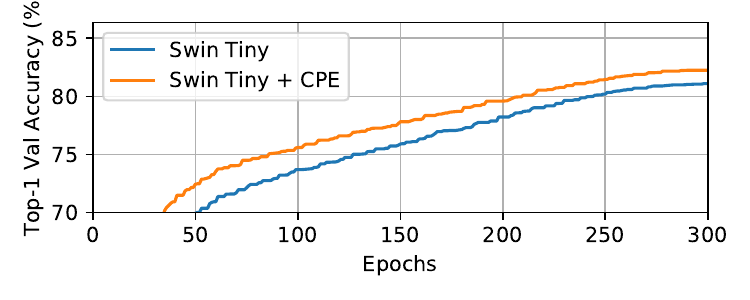}
\caption{CPE boosts Swin Tiny on ImageNet by 1.15\% top-1 Acc. }
\label{fig:cpvt-swin}
\end{figure}

We show the validation curves when training Swin \citep{liu2021swin} equipped with PEG in Figure~\ref{fig:cpvt-swin}. It can boost Swin-tiny from 81.10\% to 82.25\% (+1.15\%$\uparrow$) on ImageNet.

\subsection{Evaluation on Segmentation and Detection}\label{app:seg-det}

\paragraph{Semantic segmentation on ADE20K.}
We evaluate the performance of PEG on  the ADE20K \citep{zhou2017scene} segmentation task. Based on the Semantic FPN framework \citep{kirillov2019panoptic},  PVT achieves much better results than ResNet \citep{he2016deep} baselines. Under carefully controlled settings,  PEG further boosts PVT-tiny by 3.1\% mIoU.

\paragraph{Object detection on COCO.} We also perform controlled experiments with the RetinaNet \citep{lin2017focal} framework on the COCO detection task. The results are shown in Table~\ref{tab: PVT_peg}. In the standard 1$\times$ schedule, PEG improves PVT-tiny by 2.0\% mAP. PEG brings 2.4\% higher mAP under the 3 $\times$ schedule.

\begin{table*}[ht]
	\caption{ Our method boosts the performance of PVT on ImageNet classification, ADE20K segmentation and COCO detection}
	\label{tab: PVT_peg}
 \setlength{\tabcolsep}{1pt}
	\begin{center}		
		\small
		\begin{tabular}{l|c|c|c|c|c|c|c}
			\hline
			\multirow{2 }{*}{Backbone}& \multicolumn{2}{c|}{ImageNet}& \multicolumn{2}{c|}{Semantic FPN on ADE20K}& \multicolumn{3}{c }{RetinaNet  on COCO }  \\ 
			\cline{2-8}
			&Params & Top-1 & Params &mIoU& Params& mAP &  mAP   \\
			& (M) & (\%) & (M) & (\%)  & (M) &  (\%, 1$\times$)&  (\%, 3$\times$, +MS)  \\
			\hline 
			ResNet-18 \citep{he2016deep}&12 &69.8&16 &32.9& 21& 31.8&35.4\\
			PVT-tiny~\citep{wang2021pyramid}& 13& 75.0 &17&35.7&23&36.7&39.4 \\
		    PVT-tiny+PEG & 13 & 77.3 & 17 & 38.0 & 23& 38.0 & \textbf{41.8}\\
			PVT-tiny+GAP & 13 & 75.9 & 17 & 36.0 & 23& 36.9 & 39.7 \\
			PVT-tiny+PEG+GAP &  13& \textbf{78.1}& 17& \textbf{38.8}&23& \textbf{38.7}&\textbf{41.8}  \\
			\hline
			PVT-small~\citep{wang2021pyramid}& 25& 79.8 &28&39.8&34&40.4&42.2 \\
			PVT-small+PEG+GAP &  25& \textbf{81.2}& 28& \textbf{44.3}&34& \textbf{43.0}&\textbf{45.2}  \\
			\hline
			PVT-Medium~\citep{wang2021pyramid}& 44& 81.2 &48&41.6&54&41.9&43.2 \\
			PVT-Medium+PEG+GAP &  44& \textbf{82.7}& 48& \textbf{44.9}&54& \textbf{44.3}&\textbf{46.4}  \\
			\hline
						
		\end{tabular}
	\end{center}
\end{table*}
\subsection{Ablation on Other Forms of PEG }
We explore several forms of PEG based on the tiny model, which change the type of convolution, kernel size and layers. The inserted position  is 0.  The result is shown in Table~\ref{tab:ablation_peg_form}.  When we use large kernel of 7$\times$7 or dense convolution, the performance improvement is limited. Stacking more layers of depth-wise convolution doesn't bring significant improvement. Therefore, we use the simplest form as our default implementation. It indicates that this design  is enough to provide good position information. 

\begin{table}[ht]
	\caption{Other forms of PEG. The simple form of a single depth-wise 3$\times$3 is good enough.}
	\label{tab:ablation_peg_form}
	\setlength{\tabcolsep}{3pt}
	\begin{center}		
		\small
		\begin{tabular}{r|c|c }
			\hline
			Variants & Model & Top-1 Acc (\%)  \\

			\hline 
			1 Depthwise Conv 3$\times$3  & tiny & 72.4  \\
			1 Depthwise Conv 7$\times$7  & tiny & 72.5  \\
			4 * (Depthwise Conv 3$\times$3 +BN+ReLU)  & tiny & 72.4  \\
			1 Dense Conv 3$\times$3 & tiny & 72.3  \\
			4 * (Dense Conv 3$\times$3+BN+ReLU) & tiny & 72.5    \\

			\hline
		\end{tabular}
	\end{center}
\end{table}
\section{Example Code}

\subsection{\peg}\label{supp:peg-code}

In the simplest form, we use a single depth-wise convolution and show its usage in Transformer by the following PyTorch snippet. Through experiments, we find that such a simple design (\ie, depth-wise 3$\times$3) readily achieves on par or even better performance than the recent SOTAs. We give the torch implementation example in Alg.~\ref{alg:peg}.

\begin{algorithm}[ht]
\caption{PyTorch snippet of PEG.}
\label{alg:peg}
\definecolor{codeblue}{rgb}{0.25,0.5,0.5}
\lstset{
  backgroundcolor=\color{white},
  basicstyle=\fontsize{7.2pt}{7.2pt}\ttfamily\selectfont,
  columns=fullflexible,
  breaklines=true,
  captionpos=b,
  commentstyle=\fontsize{7.2pt}{7.2pt}\color{codeblue},
  keywordstyle=\fontsize{7.2pt}{7.2pt}\color{blue},
}
\begin{lstlisting}[language=python]
import torch
import torch.nn as nn
class VisionTransformer:
  def __init__(layers=12, dim=192, nhead=3, img_size=224, patch_size=16):
    self.pos_block = PEG(dim)
    self.blocks = nn.ModuleList([TransformerEncoderLayer(dim, nhead, dim*4) for _ in range(layers)])
    self.patch_embed = PatchEmbed(img_size, patch_size, dim*4)
  def forward_features(self, x):
    B, C, H, W = x.shape
    x, patch_size = self.patch_embed(x)
    _H, _W = H // patch_size, W // patch_size
    x = torch.cat((self.cls_tokens, x), dim=1)
    for i, blk in enumerate(self.blocks):
      x = blk(x)
      if i == 0: 
        x = self.pos_block(x, _H, _W)
    return x[:, 0]


class PEG(nn.Module):
  def __init__(self, dim=2\textsc{56}, k=3):
    self.pos = nn.Conv2d(dim, dim, k, 1, k//2, groups=dim) # Only for demo use, more complicated functions are effective too.
  def forward(self, x, H, W):
    B, N, C = x.shape
    cls_token, feat_tokens = x[:, 0], x[:, 1:]
    feat_tokens = feat_tokens.transpose(1, 2).view(B, C, H, W)
    x = self.pos(feat_tokens) + feat_tokens
    x = x.flatten(2).transpose(1, 2)
    x = torch.cat((cls_token.unsqueeze(1), x), dim=1)
    return x
\end{lstlisting}
\end{algorithm}

\section{More discussions}

\subsection{Why RPE works less well than absolute PE?}\label{app:rpe-vs-pe}

As mentioned in Section~\ref{subsec:enc-type} (main text), RPE is inferior to the absolute positional encoding. It is because RPE does not encode any absolute position information. Also discussed in Section~\ref{subsec:zero-pad} (main text), absolute position information is also important even for ImageNet classification as it is needed to determine which object is at the center of the image.  Note
that there might be multiple objects in an image, and the label of an image is the category of the object at the center. 

Additionally, although RPE becomes popular recently, it is often jointly used with absolute positional encodings (e.g., in ConViT \citep{d2021convit}), or the absolute position information is leaked in other ways (e.g., convolution paddings in CoAtNet \citep{dai2021coatnet}). This further suggests absolute position information is crucial. 

\subsection{Comparison to Lambda Networks}
Our work is also related to Lambda Networks \citep{bello2021lambdanetworks} which uses 2D relative positional encodings. We evaluate its lambda module with an embedding size of 128, where we denote its encoding scheme as RPE2D-d128. Noticeably, this configuration has about 5.9M parameters (comparable to DeiT-tiny) but only obtains 68.7\%. We attribute its failure to the limited ability in capturing the correct positional information. After all, lambda layers are designed with the help of many CNN backbones components such as down-sampling to form various stages, to replace ordinary convolutions in ResNet \citep{he2016deep}. In contrast, \ours is transformer-based.
\subsection{Qualitative Analysis of \ours}
Thus far, we have shown that PEG can have better performance than the original positional encodings. However, because PEG provides the position in an implicit way, it is interesting to see if PEG can indeed provide the position information as the original positional encodings. Here we investigate this by visualizing the attention weights of the transformers. Specifically, given a 224$\times$224 image (i.e. 14$\times$14 patches), the score matrix within a single head is 196$\times$196. We visualize the normalized self-attention score matrix of the second encoder block.

\begin{figure}[ht]
	\centering
	\includegraphics[width=0.25\columnwidth]{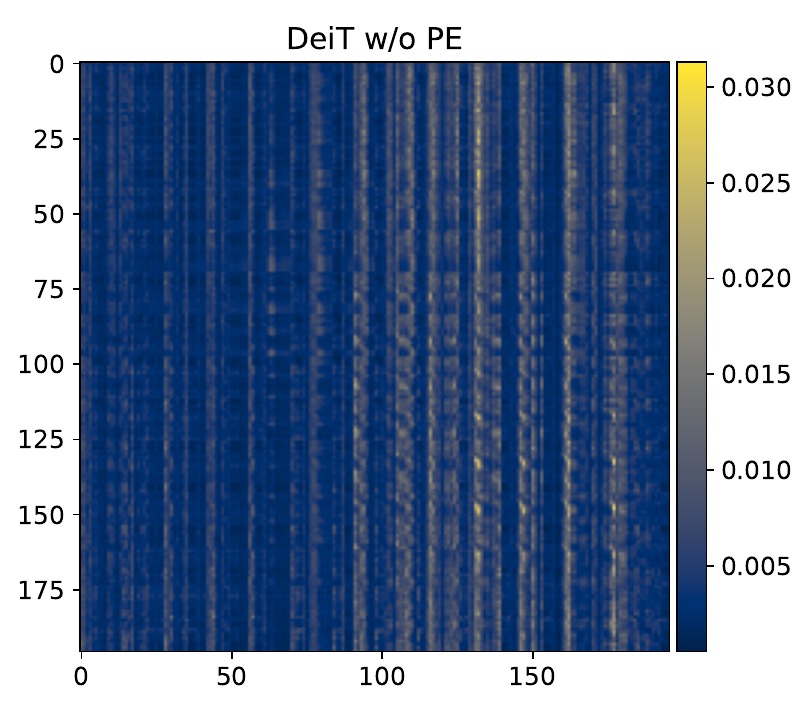}
	\includegraphics[width=0.25\columnwidth]{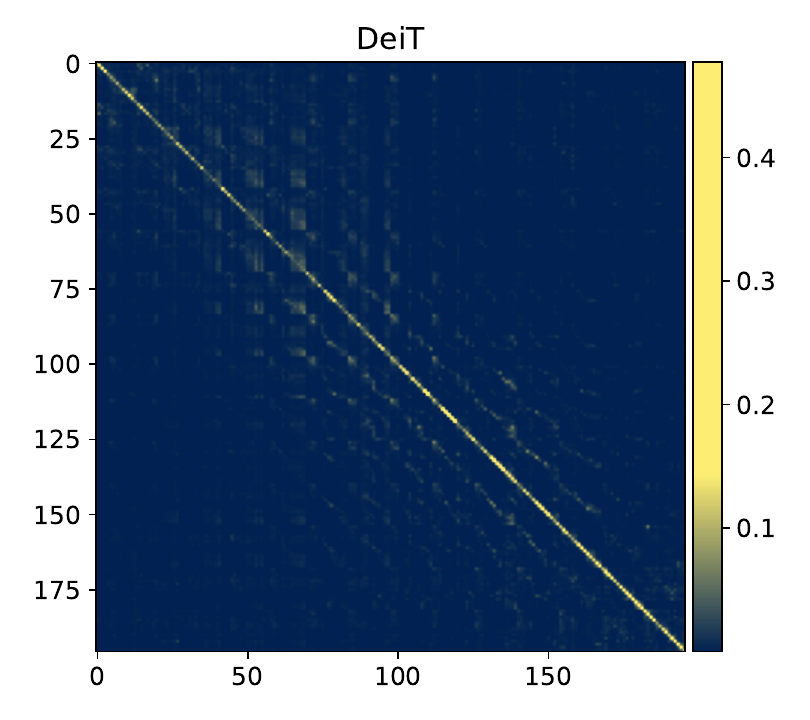}
	\includegraphics[width=0.25\columnwidth]{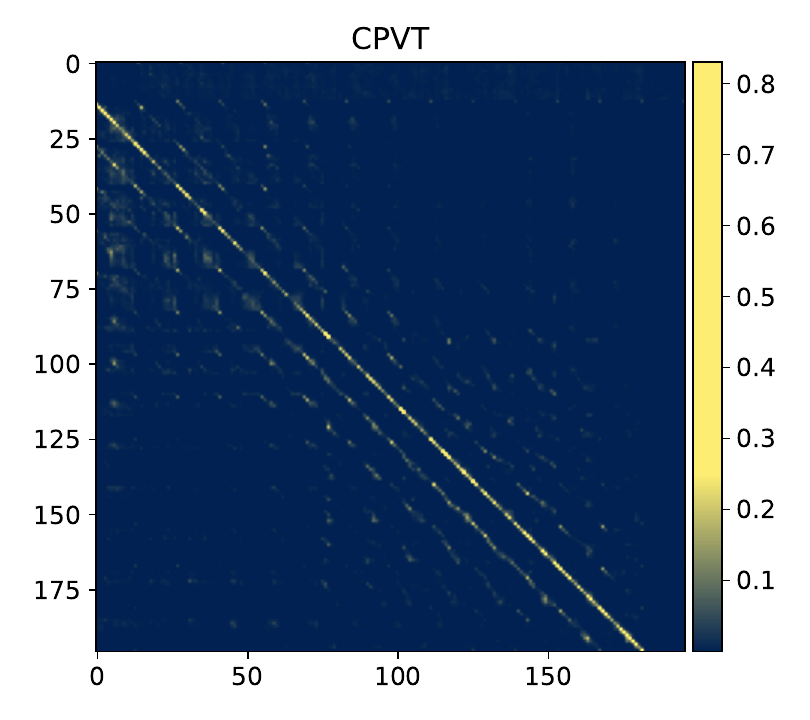}
	\caption{Normalized attention scores (first head) of the second encoder block of DeiT without position encoding  (\npe), DeiT \citep{touvron2020training}, and CPVT on the same input sequence. Position encodings are key to developing a schema of locality in lower layers of DeiT. Meantime, CPVT profits from conditional encodings and follows a similar locality pattern.}
	\label{fig:deit-cpvt-pos-comp}
\end{figure}

We first visualize the attention weights of DeiT with the original positional encodings. As shown in Figure~\ref{fig:deit-cpvt-pos-comp} (middle), the diagonal element interacts strongly with its local neighbors but weakly with those far-away elements, which suggests that DeiT with the original positional encodings learn to attend the local neighbors of each patch. After the positional encodings are removed (denoted by \npe), all the patches produce similar attention weights and fail to attend to the patches near themselves, see Figure~\ref{fig:deit-cpvt-pos-comp} (left).

Finally, we show the attention weights of our \ours model with PEG. As shown in Figure~\ref{fig:deit-cpvt-pos-comp} (right), like the original positional encodings, the model with PEG can also learn a similar attention pattern, which indicates that the proposed PEG can provide the position information as well.

\begin{figure}[ht]
	\centering
	\begin{subfigure}{0.48\linewidth}
		\includegraphics[width=0.45\columnwidth]{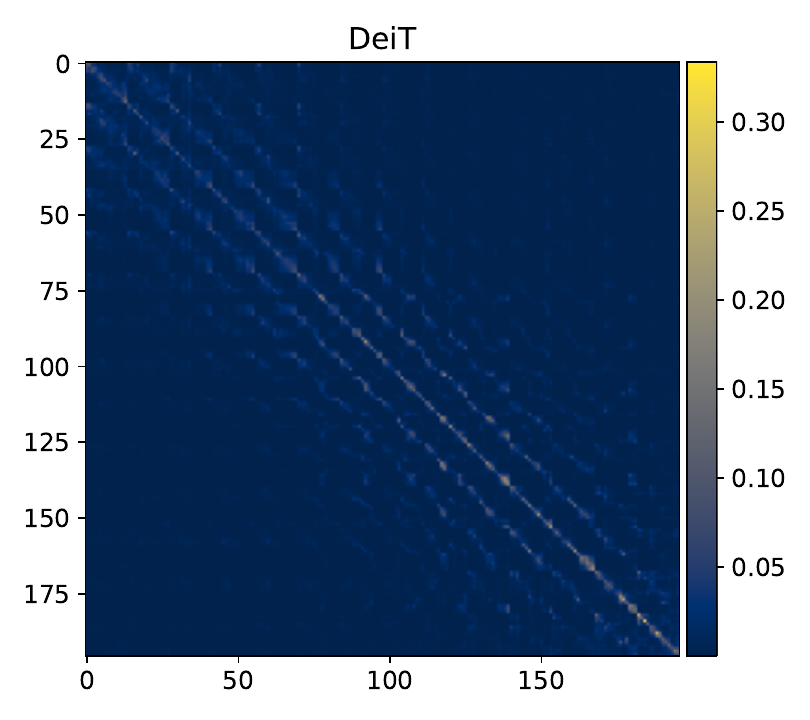}
		\includegraphics[width=0.45\columnwidth]{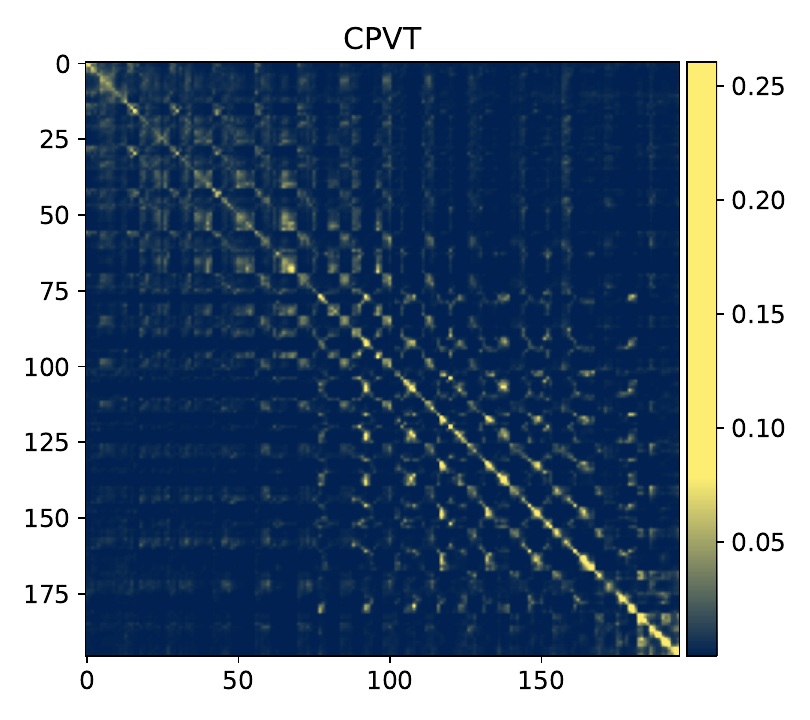}
		\caption{The second head (encoder 2)}
	\end{subfigure}
	\begin{subfigure}{0.48\linewidth}
		\includegraphics[width=0.45\columnwidth]{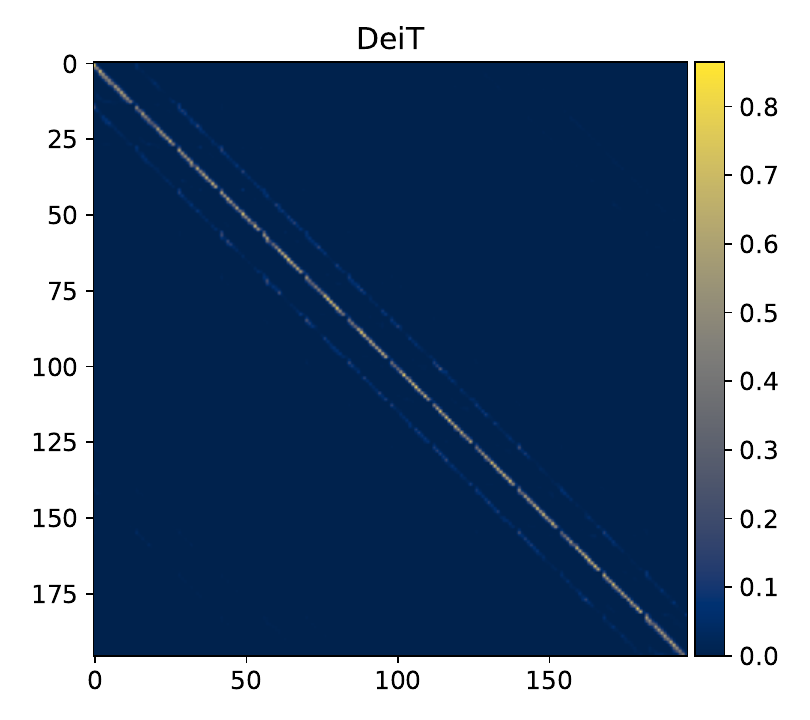}
		\includegraphics[width=0.45\columnwidth]{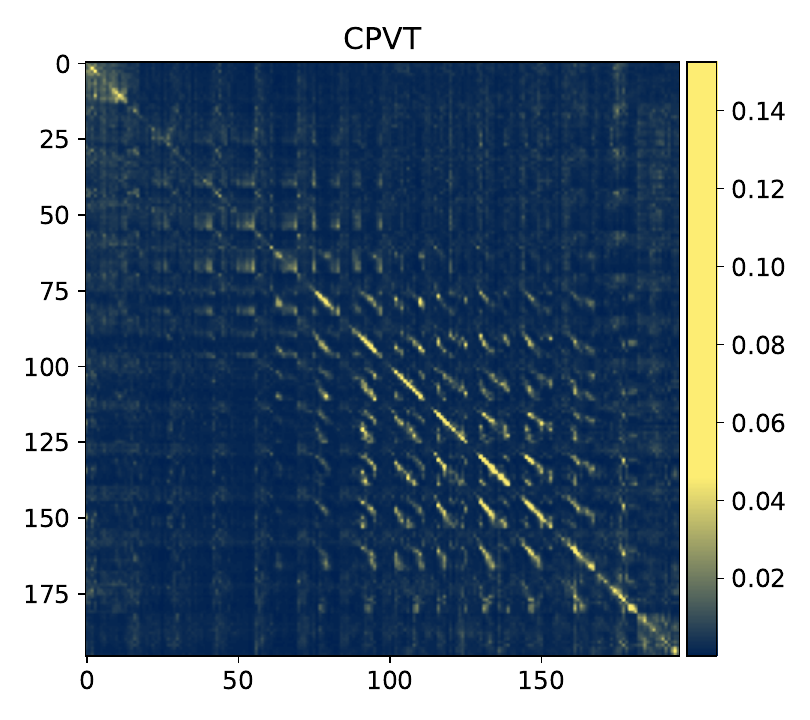}
		\caption{The third head (encoder 2)}
	\end{subfigure}
	\begin{subfigure}{0.48\linewidth}
		\includegraphics[width=0.45\columnwidth]{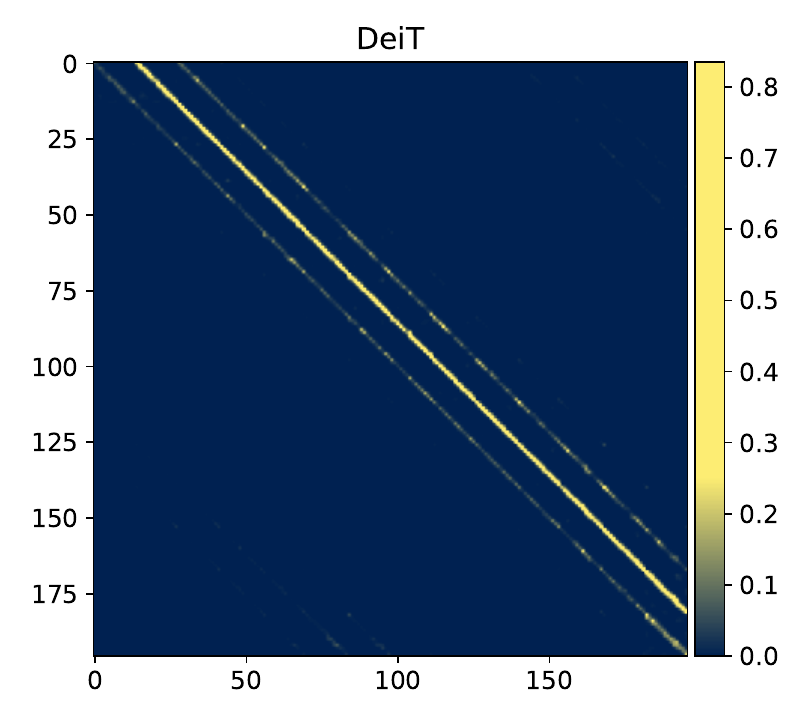}
		\includegraphics[width=0.45\columnwidth]{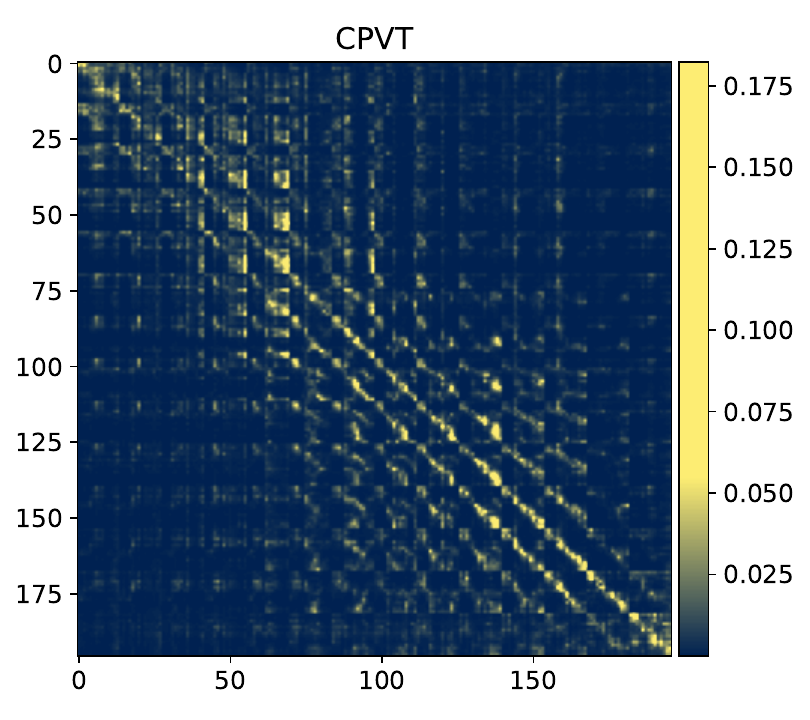}
		\caption{The first head (encoder 3)}
		\label{fig:deit-enc3-attn}
	\end{subfigure}
	\begin{subfigure}{0.48\linewidth}
		\includegraphics[width=0.45\columnwidth]{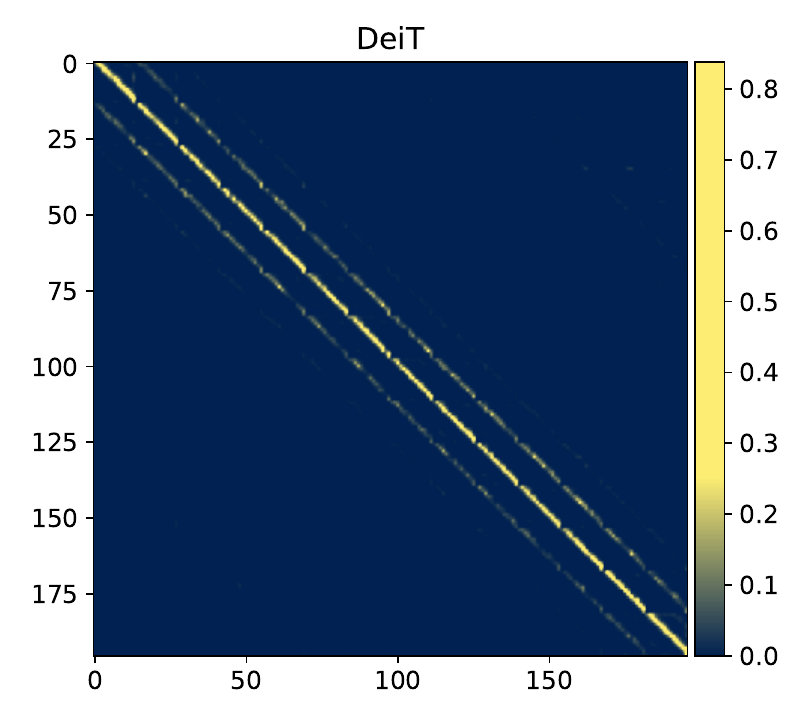}
		\includegraphics[width=0.45\columnwidth]{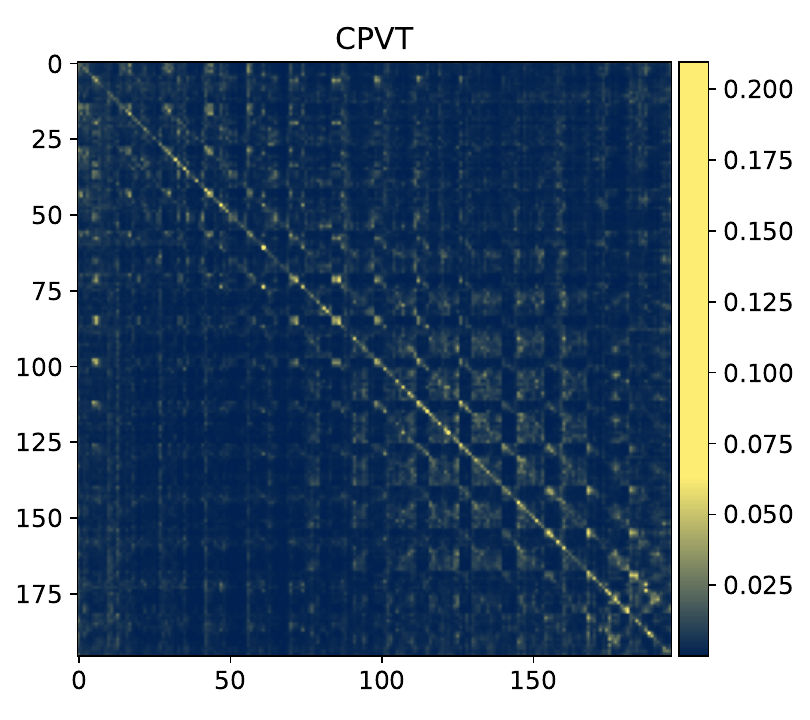}
		\caption{The second head (encoder 3)}
	\end{subfigure}
	\caption{Normalized attention scores (the second and third head) of the second and third encoder block of DeiT \citep{touvron2020training}, and CPVT on the same input sequence. DeiT and CPVT share similar locality patterns that are aligned diagonally (some might shift).}
	\label{fig:deit-cpvt-pos-comp-other-head}
\end{figure}

We illustrate the attention scores in several encoder blocks of DeiT  \citep{touvron2020training} and CPVT in the Fig.~\ref{fig:deit-cpvt-pos-comp-other-head}. It shows both methods learn similar locality patterns. As attention scores are computed over the tokens projected in different subspaces (Q and K), they do not necessarily show a strict diagonal pattern, where some may have slight shift, see DeiT in Fig.~\ref{fig:deit-enc3-attn} and CPVT of Fig.~\ref{fig:deit-cpvt-pos-comp} right.

\subsection{Comparison with Other Approaches}
We further compare our method with other approaches such as  CvT \citep{wu2021cvt}, ConViT \citep{d2021convit} and CoAtNet \citep{dai2021coatnet} on ImageNet validation set in Table~\ref{tab:comparison_with_other}. To make fair comparisons, we categorize these methods into two groups: plain and pyramid models. Since our models are primarily for plain models, we adapt our methods on two popular pyramid frameworks PVT and Swin. Our CPVT-S-GAP slightly outperforms ConViT-S by 0.2\% with 4M fewer parameters and 0.8G fewer FLOPs. When equipped with pyramid designs, our methods are still comparable to CvT and CoAtNet. 
\begin{table}[h]
	\caption{Performance comparison with other approaches such as CvT \citep{wu2021cvt}, ConViT \citep{d2021convit} and CoAtNet \citep{dai2021coatnet} on ImageNet validation set. All the models are trained on ImageNet-1k dataset and tested on the validation set using 224$\times$224 resolution.}
	\label{tab:comparison_with_other}
	\setlength{\tabcolsep}{3pt}
	\begin{center}
		\small
		\begin{tabular}{l|c|c|c|c}
			\hline
			Model & Type& Params & FLOPs & Top-1 Acc\\
			& & & &(\%)   \\
			\hline 
			DeiT-small \citep{touvron2020training} &Plain& 22M & 4.6G & 79.9\\
			ConViT-S \citep{d2021convit}&Plain&27M&5.4G&81.3\\
			CPVT-S-GAP (ours) &Plain&23M& 4.6G &\textbf{81.5}\\
			\hline
			CoAtNet-0 \citep{dai2021coatnet} &Pyramid&25M&4.2G&81.6\\
			CvT-13 \citep{wu2021cvt}&Pyramid& 20M & 4.5G& 81.6\\ 
			PVT-small \citep{wang2021pyramid} & Pyramid&25M&3.8G&79.8\\
			PVT-small+PEG+GAP & Pyramid&25M & 3.8G&81.2\\
			Swin-tiny \citep{liu2021swin} &Pyramid&29M&4.5G&81.3\\
			Swin-tiny+PEG+GAP&Pyramid & 29M &4.5G& \textbf{82.3} \\
			\hline
		\end{tabular}
		\footnotesize
	\end{center}

\end{table}

\paragraph{Comparison with DeiT w/ Convolutional Projection.} Note CvT uses a depth-wise convolution in $q$-$k$-$v$ projection which they call it \emph{Convolutional Projection}. Instead of using it in all layers, we put only one of such design into DeiT-tiny and train such a model from scratch under strictly controlled settings. We insert it in the position 0 as in our method. The result is shown in Table~\ref{tab:comparison_with_cvt}.  This CvT-flavored  DeiT achieves 70.6\% top-1 accuracy on ImageNet validation set, which is lower than ours (72.4\%). Note that $q$-$k$-$v$ projections in CvT utilize three depthwise convolutions, therefore, this setting has more parameters than ours. This attests the difference of CvT and CPVT, verifying our advantage by learning better position encodings other than inserting them in all layers to have the ability to capture local context and to remove ambiguity in attention.

\begin{table}[h]
	\caption{Comparison with positional encoding in CvT \citep{wu2021cvt} on ImageNet validation set. All the models are trained on ImageNet-1k dataset and tested on the validation set using 224$\times$224 resolution.}
	\label{tab:comparison_with_cvt}
	\setlength{\tabcolsep}{3pt}
	\begin{center}
		\small
		\begin{tabular}{ l|c|c|c}
			\hline
			Model & Params &Insert Position & Top-1 Acc\\
			& & &(\%)   \\
			\hline 
				\ourstiny &   5681320&0& 72.4 \\
				DeiT+ Convolutional Projection &5685352&0&70.6\\
			\hline
		\end{tabular}
		\footnotesize
	\end{center}
	
\end{table}

\end{document}